\documentclass[10pt,twocolumn,letterpaper]{article}

\usepackage{cvpr}              % To produce the CAMERA-READY version
\usepackage[accsupp]{axessibility}
\definecolor{cvprblue}{rgb}{0.21,0.49,0.74}
\usepackage[pagebackref,breaklinks,colorlinks,allcolors=cvprblue]{hyperref}

\usepackage{algorithm}
\usepackage{algorithmicx}
\usepackage{amsmath}
\usepackage{algorithm}
\usepackage{algpseudocode}
\usepackage{multirow}

\usepackage{xcolor} % Required for colors
\usepackage{listings}

\definecolor{codegreen}{rgb}{0,0.6,0}
\definecolor{codegray}{rgb}{0.5,0.5,0.5}
\definecolor{codeblue}{rgb}{0,0,0.8}
\definecolor{codered}{rgb}{0.8,0,0}

\lstdefinestyle{json}{
    language=C, % Use C as a fallback, it's almost always included
    basicstyle=\ttfamily\small,
    otherkeywords={true,false,null}, % JSON keywords
    keywordstyle=\color{codeblue}, % Keyword color
    stringstyle=\color{codered}, % String color
    commentstyle=\color{codegreen},
    ndkeywords={prompt, targets}, % Highlight your specific keys
    ndkeywordstyle=\color{codeblue}\bfseries, % Make your keys bold blue
    breakatwhitespace=false,
    breaklines=true,
    captionpos=b,
    keepspaces=true,
    numbers=left,
    numbersep=5pt,
    numberstyle=\tiny\color{codegray},
    showspaces=false,
    showstringspaces=false,
    showtabs=false,
    tabsize=2,
    frame=tb, % Adds a top and bottom frame
    backgroundcolor=\color{white} % Set background color
}

\def\paperID{12417} % *** Enter the Paper ID here
\def\confName{CVPR}
\def\confYear{2026}

\title{TokenTrace: Multi-Concept Attribution through Watermarked Token Recovery}

\author{Li Zhang$^{1,2}$ \\
\and Shruti Agarwal$^{1}$ \\
\and John Collomosse$^{1,3}$\\
\and Pengtao Xie$^{2}$
\and  Vishal Asnani$^{1}$\\
\and \vspace{-1.1cm}\\
$^{1}$Adobe Research, $^{2}$University of California, San Diego,
$^{3}$DECaDE, University of Surrey\\
{\tt\small  \{liz042, p1xie\}@ucsd.edu ~~ \{shragarw, collomos, vasnani\}@adobe.com}
}

\begin{document}

\maketitle
\begin{abstract}
Generative AI models pose a significant challenge to intellectual property (IP), as they can replicate unique artistic styles and concepts without attribution. While watermarking offers a potential solution, existing methods often fail in complex scenarios where multiple concepts (e.g., an object and an artistic style) are composed within a single image. These methods struggle to disentangle and attribute each concept individually. In this work, we introduce TokenTrace, a novel proactive watermarking framework for robust, multi-concept attribution. Our method embeds secret signatures into the semantic domain by simultaneously perturbing the text prompt embedding and the initial latent noise that guide the diffusion model's generation process. For retrieval, we propose a query-based TokenTrace module that takes the generated image and a textual query specifying which concepts need to be retrieved (e.g., a specific object or style) as inputs. This query-based mechanism allows the module to disentangle and independently verify the presence of multiple concepts from a single generated image. Extensive experiments show that our method achieves state-of-the-art performance on both single-concept (object and style) and multi-concept attribution tasks, significantly outperforming existing baselines while maintaining high visual quality and robustness to common transformations.
\end{abstract}    
\section{Introduction}
\label{sec:intro}

Text-to-image foundation models~\cite{ruiz2023dreambooth,kim2024transparent,awais2025foundation}, propelled by large-scale diffusion architectures~\cite{rombach2022high,esser2024scaling}, have demonstrated unprecedented success in generating high-fidelity, complex visual content from natural language descriptions. This progress has brought generative AI to the forefront of creative industries~\cite{huang2025diffusion}. However, this power introduces a critical challenge in proactive attribution: the need to embed imperceptible, robust watermarks into generated content to verify ownership and trace provenance~\cite{fernandez2023stable,zhu2024watermark}. This task is fundamental to protecting the intellectual property of artists and creators whose unique styles and concepts are used to train these models. Recent state-of-the-art methods, such as ProMark~\cite{asnani2024promark}, have pioneered a ``proactive'' approach, embedding imperceptible watermarks directly into the pixel space of images. These methods train the diffusion model to preserve these spatial watermarks, allowing a decoder to later detect them in generated images and establish a causal connection to the original training examples that contribute to the generations. Such invention is effective for the general concept attribution task.

Despite progress in this area, existing watermarking methods face notable limitations. Approaches that operate in the pixel domain are often fragile, failing to survive common transformations like compression or cropping~\cite{zhu2024watermark,asnani2024promark,sander2025watermark}. While more recent methods embedding watermarks into the latent space offer greater robustness~\cite{rezaei2024lawa,wang2025sleepermark,asnani2025custom}, they fail to address a significant challenge posed by the compositional nature of generative AI. These models frequently generate images by combining multiple, distinct concepts (e.g., a specific character rendered in a unique artistic style). Existing attribution methods, which embed a single holistic watermark, are not designed for this scenario; they cannot disentangle and independently attribute multiple concepts when their visual representations overlap in the final image. This failure to attribute composed concepts is a major gap in protecting creator IP. While some recent works attempt multi-concept attribution~\cite{wang2024must}, they struggle with signal interference and lack a precise mechanism for targeted retrieval.

\begin{figure}
    \centering
    \includegraphics[width=1\linewidth]{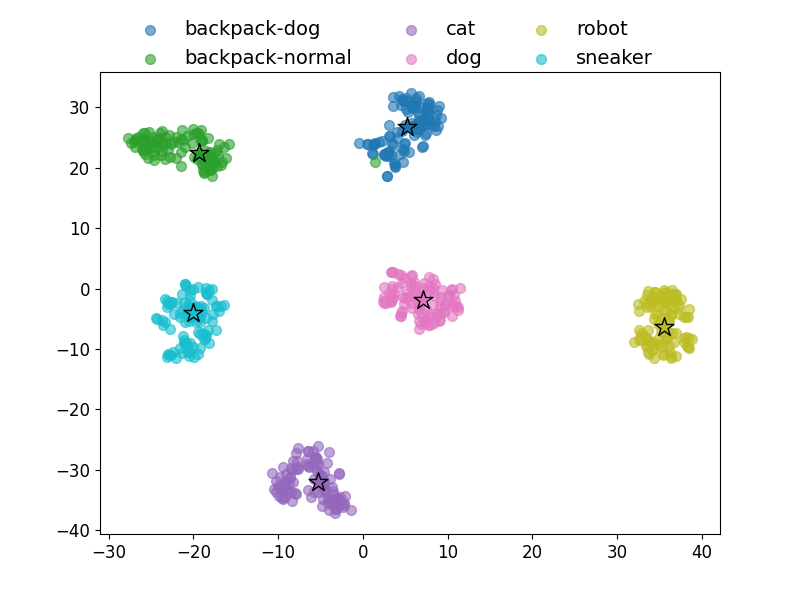}
    \vspace{-1cm}
    \caption{t-SNE visualization of predicted concept embeddings. The embeddings retrieved by TokenTrace module from DreamBooth-generated images (dots) form distinct, well-separated clusters around their ground-truth embeddings (stars), validating its ability to perform concept attribution.}
    \vspace{-0.5cm}
    \label{fig:tsne}
\end{figure}

In response to the above challenges, we introduce TokenTrace, a proactive watermarking framework for robust, multi-concept attribution. We hypothesize that a watermark's robustness and specificity can be dramatically improved by associating it directly with the textual semantics of the concept it represents, a strategy inspired by the success of prompt-tuning in foundation models~\cite{zhou2022conditional,jia2022visual}. This association is the key to solving the multi-concept challenge: instead of embedding competing signals into the limited pixel space, we link each concept's secret to its unique representation within the text embedding. This approach fundamentally avoids the problem of spatial overlap, as the signatures are separated in the textual semantic domain before generation begins. TokenTrace achieves this via a novel dual-conditioning strategy. A concept's secret is processed by two parallel networks: a concept encoder that perturbs the text prompt embedding and a secret mapper that perturbs the initial latent noise. By embedding the signature in both the semantic and latent domains, the watermark is deeply integrated into the generative process, ensuring both robustness and this crucial concept-level separation.
To retrieve the watermark, we propose a query-based TokenTrace module designed to reverse this watermark encoding, which takes the watermarked image and a query prompt as inputs to predict the corresponding concept secrets. The feasibility of this retrieval approach was validated in a preliminary experiment~\footnote{Detailed settings for this preliminary experiment can be found in the supplementary material.}; as shown in a t-SNE visualization (Figure~\ref{fig:tsne}), TokenTrace module successfully maps generated images back to distinct, well-separated clusters for each concept.

%% MOVE TO SEC 3
% To retrieve the watermark, we propose a query-based TokenTrace module designed to reverse this watermark encoding. This module takes the watermarked image and a query prompt (e.g., "a photo of $<$sks-duck$>$") as input. The query is essential, as it provides the context for the module to isolate the specific concept signature it needs to retrieve. It then uses a multi-modal attention mechanism to predict the corresponding concept embedding, which a simple decoder translates back into the original secret. 
% We validated this premise in a preliminary experiment where we trained the TokenTrace module on real DreamBooth images and used it to predict embeddings from generated images.~\footnote{Detailed settings for this preliminary experiment can be found in the supplementary material.} As shown in a t-SNE visualization (Figure~\ref{fig:tsne}), the predicted embeddings form distinct, well-separated clusters, confirming that our retrieval approach is feasible.
% By segregating the retrieval process based on a user's textual query, our method can effectively disentangle and independently verify multiple, composed concepts from a single image, thus addressing the spatial overlap problem that plagues pixel-based methods.
%%

\noindent Our work makes three key contributions:
\begin{itemize}
    \item We propose TokenTrace, a novel framework that embeds watermarks into both the text prompt and latent noise domains, intrinsically tying the watermark to the concept's semantics.

    \item We introduce a query-based module that can effectively disentangle and independently attribute multiple, overlapping concepts (such as objects and styles) from a single generated image.

    \item Our extensive experiments on both object and style concepts demonstrate that TokenTrace significantly outperforms state-of-the-art attribution baselines in both single-concept and multi-concept scenarios, all while maintaining high visual fidelity.
\end{itemize}

\section{Related Works}
\label{sec:related}

\paragraph{Generative Customization}
The advent of large-scale text-to-image diffusion models has enabled remarkable image synthesis~\cite{saharia2022photorealistic,rombach2022high,ruiz2023dreambooth,zhang2024forget,wang2025designdiffusion}. A significant and parallel line of research is generative customization, which focuses on personalizing these foundation models with novel, user-provided concepts~\cite{kumari2023multi,wei2023elite,chen2024anydoor,li2024photomaker,zhang2025easycontrol}. Techniques such as Textual Inversion~\cite{gal2022image} and DreamBooth~\cite{ruiz2023dreambooth} allow models to learn a new concept from just a few example images, often by learning a new pseudo-token in the model's text-embedding space or leaning a new set of model parameters for a customized concept. While this customization empowers creators to inject their unique intellectual property (IP) into the generative process, it also exposes that IP to misuse~\cite{thongmeensuk2024rethinking,tyagi2024copyright,chesterman2025good}. These learned concepts can be extracted and repurposed without permission, which introduces a critical need for robust attribution methods designed specifically for these personalized concepts.

\vspace{-0.3cm}
\paragraph{Watermarking and Concept Attribution for Generative Models} To address provenance and attribution, various watermarking techniques for generative models have been proposed~\cite{fernandez2023stable,yang2024gaussian,huang2024robin,wang2025sleepermark,muller2025black}, which are broadly categorized as passive or proactive. Passive watermarking applies a signature after generation~\cite{ganic2004robust,singh2012novel,zhu2018hidden}. This approach, while simple, adds computational cost, is easily defeated by common transformations, and is thus insufficient for reliable attribution.
In contrast, proactive watermarking, our focus, integrates the signature during the generative process~\cite{asnani2024promark,huang2024robin,wang2025roar}, making it inherently more robust to transformations and the necessary choice for causal attribution. Existing proactive methods embed signatures in the pixel-domain~\cite{asnani2024promark,sander2025watermark} or latent-domain~\cite{yang2024gaussian,asnani2025custom}. However, these traditional methods typically embed a single, content-agnostic watermark, failing to link it to specific semantic concepts, which leads to the challenges we address next.

\vspace{-0.3cm}
\paragraph{Proactive Schemes} Proactive schemes embed signals into the generative process for tasks like deepfake detection~\cite{wang2021faketagger,zhao2023proactive}, manipulated content localization~\cite{asnani2022proactive,zhang2025omniguard, asnani2023malp}, and concept attribution~\cite{asnani2024promark,asnani2025custom}. While pixel-domain methods are fragile and fail when multiple concept spatially overlap, and latent-domain methods are robust but content-agnostic, a new category operates in the semantic domain~\cite{nguyen2025survey}, linking watermarks to textual representations. For instance, ``Guarding Textual Inversion"~\cite{feng2023catch} is designed for single-concept tracing and does not address compositional attribution. CustomMark~\cite{asnani2025custom} also customizes models via text prompts for multi-concept attribution but struggles to disentangle concepts that are spatially overlapping. Thus, a robust, query-based mechanism for disentangling composed concepts remains a significant and unaddressed challenge.
% \paragraph{Proactive Schemes}
% Proactive schemes enhance various tasks for generative models by embedding signals or perturbations into the generative process, providing benefits, including deepfake detection~\cite{wang2021faketagger,zhao2023proactive}, manipulated content localization~\cite{asnani2022proactive,zhang2025omniguard}, and concept attribution~\cite{asnani2024promark,asnani2025custom}. Some approaches focus on embedding these signals in the pixel domain; however, these are often fragile and fail when multiple concept-signals spatially overlap. Other methods embed watermarks into the latent space, offering greater robustness but remaining content-agnostic, as the signal is not tied to the image's semantic meaning.
% To address these limitations, a more recent category of proactive schemes operates in the semantic domain~\cite{nguyen2025survey}. These methods link the watermark to the concept's textual representation.
% For instance, "Guarding Textual Inversion"~\cite{feng2023catch} embeds a watermark directly into a concept's token embedding, but its framework is designed for single-concept tracing and does not address compositional attribution. "CustomMark"~\cite{asnani2025custom} is a more direct baseline that also customizes a model to embed watermarks via text prompts and claims to support multi-concept attribution. However, a robust, query-based mechanism for disentangling these composed concepts remains a challenge.

\begin{figure*}
    \centering
    \includegraphics[width=1\linewidth]{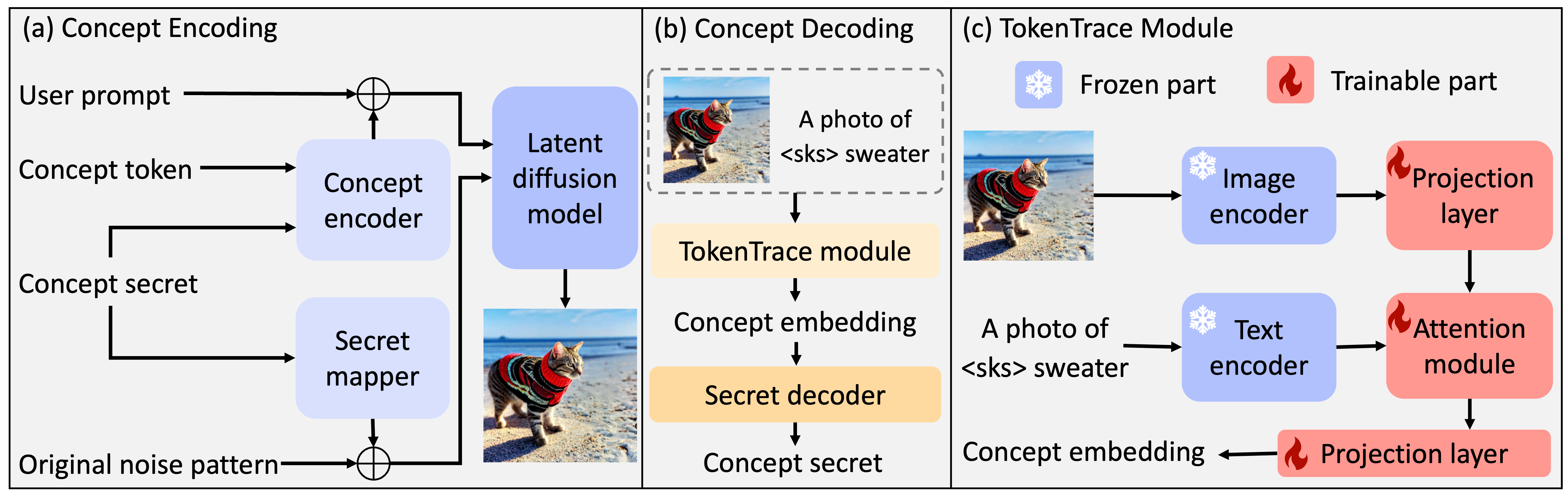}
    \vspace{-0.7cm}
    \caption{Overview of TokenTrace. (a) Concept encoding: A concept secret is fed into a concept encoder to perturb the targeted concept token and a secret mapper to perturb the initial noise, enabling a dual-conditioning of the latent diffusion model. (b) Concept decoding: The generated image and a query prompt are fed into the TokenTrace module to predict a concept embedding, which a secret decoder then translates back into the original concept secret for verification. (c) TokenTrace module: We use frozen image and text encoders, whose features are fused by trainable projection and attention layers to predict the final concept embedding.}
    \vspace{-0.5cm}
    \label{fig:method_overview}
\end{figure*}

\section{Method}
\label{sec:method}

In this section, we present our method, TokenTrace, a proactive watermarking framework designed for robust, multi-concept attribution that can be used as a controlled-generation provenance tool. As illustrated in Figure~\ref{fig:method_overview}, our framework is composed of two stages: (a) a concept encoding stage that embeds concept secrets into the generative process, (b) a concept decoding stage that retrieves the concept secrets that is indicated by a query prompt, and (c) the detailed architecture of our TokenTrace module.

\subsection{Concept Encoding}
Our encoding process employs a novel dual-conditioning strategy that embeds a watermark by simultaneously perturbing the generative inputs in both the semantic domain (textual embeddings) and the latent domain (initial noise). As shown in Figure~\ref{fig:method_overview}(a), this stage involves two parallel networks that process a concept secret ($\mathcal{S}$):
\begin{itemize}
    \item The concept encoder ($f_{\text{enc}}$) targets the specific concept token to be watermarked. A user prompt embedding $E_{\text{prompt}}$ is a collection of token embeddings $\{e_\text{1}, \dots, e_\text{c}, \dots, e_\text{k}\}$, where $e_\text{c}$ is the target concept token embedding. The encoder $f_{\text{enc}}$ takes the concept secret ($\mathcal{S}$) and the corresponding concept token embedding ($e_\text{c}$) as input to generate a perturbation. This perturbation is fused (via element-wise addition) only with the target token embedding $e_\text{c}$ to create a perturbed token embedding $\hat{e}_\text{c}$. The final perturbed prompt embedding, $\hat{E}_{\text{prompt}}$, is calculated by replacing the original token:
    \begin{equation}
    \left\{
    \begin{aligned}
    &\hat{e}_\text{c} = e_\text{c} + f_{\text{enc}}(e_\text{c}, \mathcal{S}) \\
    &\hat{E}_{\text{prompt}} = \{e_\text{1}, \dots, \hat{e}_\text{c}, \dots, e_\text{k}\}
    \end{aligned}
    \right.
    \end{equation}

    % \item The concept encoder ($f_{enc}$) takes both the concept secret ($\mathcal{S}$) and a corresponding concept prompt ($E_{concept}$) as input. It outputs a perturbation embedding which is fused (via element-wise addition) with the user prompt embedding ($E_{prompt}$) to create the final perturbed text embedding, $\hat{E}_{prompt}$.
    % \begin{align}
    %     \hat{E}_{prompt} = E_{prompt} + f_{enc}(E_{prompt}, \mathcal{S})
    % \end{align}
    
    \item The secret mapper ($f_{\text{map}}$) takes only the concept secret ($\mathcal{S}$) as input and generates a structured perturb gaussian pattern. This pattern is fused with the original noise pattern ($z_\text{T}$) to create the perturbed initial noise,
    $\hat{z}_\text{T}$.
    \begin{align}
        \hat{z}_\text{T} = z_\text{T} + f_{\text{map}}(\mathcal{S})
    \end{align}
\end{itemize}
Finally, the diffusion model ($DM$) is conditioned on both of these inputs, the prompt embedding with perturbed concept token embeddings ($\hat{E}_{\text{prompt}}$) and the perturbed noise ($\hat{z}_\text{T}$), to generate the final watermarked image ($I_{\text{wm}}$).
\begin{align}
    I_{\text{wm}} = DM(\hat{z}_\text{T}, \hat{E}_{\text{prompt}})
\end{align}

By applying this dual-conditioning strategy, our method achieves deep integration. Specifically, by embedding the signature in both the semantic and latent domains, the watermark is fundamentally woven into the image's structure, making it far more robust to transformations than methods that only modify pixel space.

\subsection{Concept Decoding} \label{sec:method_decode}

As shown in Figure~\ref{fig:method_overview}(b), the decoding stage is a two-step pipeline designed to reliably retrieve the embedded watermark and confirm its origin. First, the watermarked image ($I_{\text{wm}}$) and a textual query ($P_{\text{query}}$) are fed into the TokenTrace module ($f_{\text{tt}}$). This textual query is a simple prompt, selected from a pre-defined set, that specifies which concept to retrieve. For example, to retrieve the $<$sks-object$>$ concept, the query $P_{\text{query}}$ would be ``a photo of $<$sks-object$>$". The TokenTrace module then uses both inputs to predict the concept embedding ($\tilde{e}_{\text{c}}$).
\begin{align}
    \tilde{e}_{\text{c}} = f_{\text{tt}}(I_{\text{wm}}, P_{\text{query}})
\end{align}
This predicted embedding is then passed to the secret decoder ($f_{\text{dec}}$), which we implement as a simple linear network. This decoder translates the high-dimensional embedding back into the original bit-secret, $\tilde{\mathcal{S}}$.
\begin{align}
    \tilde{\mathcal{S}} = f_{\text{dec}}(\tilde{e}_{\text{c}})
\end{align}

This query-based pipeline provides a direct solution to the multi-concept challenge. By requiring a specific text query ($P_{\text{query}}$) to initiate retrieval, the system can selectively isolate and verify a single concept's signature. This mechanism allows it to successfully disentangle and attribute multiple, overlapping concepts (e.g., an object and a style) from a single image, a task that pixel-based methods cannot perform.

Importantly, the core of our retrieval mechanism is the TokenTrace module, which is architected to effectively fuse multi-modal information and predict a concept's original embedding from a generated image. As detailed in Figure~\ref{fig:method_overview}(c), the module is designed for parameter-efficient finetuning by leveraging powerful frozen pre-trained encoders from CLIP~\cite{radford2021learning} for robust feature extraction.
The data flow is as follows:
\begin{enumerate}
    \item The watermarked image $I_{\text{wm}}$ is passed through the frozen Image encoder ($f_{\text{img\_enc}}$). Its output features are then fed into a trainable projection layer ($f_{\text{proj\_1}}$) to align their dimensionality with the text features.

    \item The query prompt $P_{\text{query}}$ is passed through the frozen Text encoder ($f_{\text{text\_enc}}$).

    \item The aligned image features and the text features are fed into a trainable attention module ($f_{\text{attn}}$), which produces a fused, context-aware representation.

    \item This fused representation is passed through a final trainable projection layer ($f_{\text{proj\_2}}$) to generate the predicted concept embedding, $\tilde{e}_{\text{c}}$.
\end{enumerate}
This entire retrieval process is summarized by:
\begin{equation}
\left\{
\begin{aligned}
&F_{\text{img}} = f_{\text{proj\_1}}(f_{\text{img\_enc}}(I_{\text{wm}})) \\
&F_{\text{text}} = f_{\text{text\_enc}}(P_{\text{query}}) \\
&F_{\text{fused}} = f_{\text{attn}}(F_{\text{img}}, F_{\text{text}}) \\
&\tilde{e}_{\text{c}} = f_{\text{proj\_2}}(F_{\text{fused}})
\end{aligned}
\right.
\end{equation}

The TokenTrace module is designed for high parameter efficiency by leveraging powerful, frozen CLIP encoders for feature extraction. This design, inspired by adapter-based finetuning~\cite{houlsby2019parameter}, requires only a few lightweight projection and attention layers to be trainable. It allows the module to be adapted to new concepts quickly and scalably, avoiding the high computational cost and potential catastrophic forgetting associated with fully finetuning large-scale models.

\subsection{Training Objective}
Our training objective is to jointly optimize all trainable components: the concept encoder, secret mapper, the trainable projection and attention layers of the TokenTrace module, and the secret decoder. The model is trained using a composite loss function, $\mathcal{L}_{\text{total}}$, which is designed to simultaneously satisfy two competing goals: watermark retrieval accuracy and visual fidelity. This final loss is a weighted sum of four individual constraints:
\begin{itemize}
    \item A cross-entropy loss between the original and predicted secrets to ensure the watermark can be accurately retrieved:
    \begin{align}
        \mathcal{L}_{\text{BCE}} = \text{BCE}(\mathcal{S}, \sigma(\hat{\mathcal{S}}_{\text{logits}})),
    \end{align}
    where $\sigma$ is the sigmoid function, $\mathcal{S}$ is the ground-truth, and $\hat{\mathcal{S}}_{\text{logits}}$ is the output logits from the secret decoder.

    \item A contrastive style descriptor (CSD) loss~\cite{somepalli2024measuring} to maintain high-level semantic consistency between the watermarked and clean images:
    \begin{align}
        \mathcal{L}_{\text{CSD}} = 1 - \frac{\phi(I_{\text{clean}}) \cdot \phi(I_{\text{wm}})}{\|\phi(I_{\text{clean}})\| \|\phi(I_{\text{wm}})\|},
    \end{align}
    where $\phi$ represents the feature extractor, $I_{\text{clean}}$ is the clean image, and $I_{\text{wm}}$ is the watermarked image.

    \item A standard L2 loss to minimize perceptible, pixel-level differences between the watermarked and clean images, ensuring imperceptibility:
    \begin{align}
        \mathcal{L}_{\text{L2}} = \| I_{\text{clean}} - I_{\text{wm}} \|_2^2.
    \end{align}

    \item A regularization loss between the predicted and original concept embeddings to ensure the TokenTrace module's output is accurate:
    \begin{align}
        \mathcal{L}_{\text{reg}} = \| e_{\text{c}} - \tilde{e}_{\text{c}} \|_2^2,
    \end{align}
    where $e_{\text{c}}$ and $\tilde{e}_{\text{c}}$ are the ground truth and predicted concept embeddings.
\end{itemize}
During training, the final loss is the weighted sum of these individual loss constraints, which are jointly optimized to balance watermark retrieval accuracy with visual fidelity:
\begin{align}\label{loss_all}
    \mathcal{L}_{\text{total}} = \lambda_1 \mathcal{L}_{\text{BCE}} + \lambda_2 \mathcal{L}_{\text{CSD}} + \lambda_3 \mathcal{L}_{\text{L2}} + \lambda_4 \mathcal{L}_{\text{reg}},
\end{align}
where $\{\lambda_1, \lambda_2, \lambda_3, \lambda_4\}$ are the trade-off parameters to balance different kind of constrains.
By using this composite loss function, we can jointly optimize for two competing objectives: retrieval accuracy (via $\mathcal{L}_{\text{BCE}}$ and $\mathcal{L}_{\text{reg}}$) and visual fidelity (via $\mathcal{L}_{\text{L2}}$ and $\mathcal{L}_{\text{CSD}}$). This ensures the resulting watermark is both robustly detectable and imperceptible.

The end-to-end training procedure for TokenTrace is detailed in Algorithm 1 in the supplementary material. In summary, we jointly optimize all trainable parameters. In each training iteration, we sample a batch of data (including a clean image, prompts, secret, and ground-truth embedding) and perform the concept encoding step to generate a watermarked image. This image is then passed through the decoding pipeline to retrieve the secret. Finally, we compute the total composite loss $\mathcal{L}_{\text{total}}$ and update all trainable parameters via gradient descent. The inference (verification) process is a simple forward pass of the trained decoding modules ($f_{\text{tt}}$ and $f_{\text{dec}}$), as described in Section~\ref{sec:method_decode}.

\section{Experiments}
\label{sec:experiments}

In this section, we present a comprehensive evaluation of our proposed TokenTrace framework. We evaluate its effectiveness, scalability, and unique capability for multi-concept attribution across a diverse set of tasks: single style concept attribution, large-scale single object attribution, and compositional attribution using both customized and general concepts. We compare our framework against several state-of-the-art passive and proactive baselines and conduct a series of ablation studies to validate our key design choices.

\subsection{Datasets}
For the single style concept attribution task, we employ $23$ distinct styles sourced from the WikiArt dataset~\cite{tan2018improved}, which is a collection of digital artworks sourced from the online art encyclopedia~\footnote{https://www.wikiart.org/}. It is one of the most popular datasets for art-related computer vision tasks. For the single object concept attribution task, we use $1000$ object classes from the ImageNet dataset~\cite{deng2009imagenet}, which is a foundational, large-scale dataset in computer vision. For our primary task of customized multi-concept attribution, we source pre-trained concept embeddings from the Stable Diffusion Textual Inversion Embeddings library~\footnote{https://cyberes.github.io/stable-diffusion-textual-inversion-models/}. This allows us to construct challenging prompts that combine multiple concepts, such as a specific object and a specific style, to test our model's disentanglement capabilities. We also evaluate general multi-concept attribution, as composing multiple customized textual inversion embeddings in a single prompt often degrades image quality. To isolate our attribution mechanism's performance from this generation artifact, we use standard, non-customized concepts (e.g., ``a cat and a dog"). We construct a new evaluation benchmark by using ChatGPT~\cite{brown2020language} to generate a diverse set of prompts that involve the composition of multiple concepts. For each prompt, we also generate a list of target concepts that are presented in the prompt~\footnote{Specific examples can be found in supplementary material.}.

\subsection{Experimental Settings}
\paragraph{Baselines.} 
We compare our method against a comprehensive suite of baselines, which are divided into two categories. The first, passive attribution, includes methods like ALADIN~\cite{ruta2021aladin}, CLIP~\cite{radford2021learning}, Abc~\cite{wang2023evaluating}, SSCD~\cite{pizzi2022self}, and EKILA~\cite{balan2023ekila}, which attempt to attribute a generated image after its creation by measuring visual correlation or similarity to known training data. The second, proactive attribution, includes our primary competitors, which embed a causal signature into the generative process itself. This category is represented by ProMark~\cite{asnani2024promark}, a state-of-the-art method that embeds watermarks into the pixel domain, and CustomMark~\cite{asnani2025custom}, a highly relevant baseline that, like our method, operates in the semantic domain by modifying text prompts.

\vspace{-0.3cm}
\paragraph{Metrics.}
To evaluate our framework, we use two categories of metrics. For attribution assessment, we measure attribution accuracy, the percentage of times the predicted bit-secret exactly matches the correct ground-truth secret, and bit accuracy, the average percentage of individual bits correctly predicted. For image quality assessment, we measure the visual fidelity between the clean and watermarked images using two embedding-space metrics: the CSD Score, which is the cosine similarity between CSD descriptors to assess style preservation, and the CLIP Score, which is the cosine similarity between CLIP image embeddings to assess overall semantic similarity. 

\vspace{-0.3cm}
\paragraph{Hyper-parameters.} 
We implement TokenTrace in PyTorch. For the text-to-image latent diffusion model, we use the pre-trained Stable Diffusion 1.5~\cite{rombach2022high}. For the TokenTrace module, we utilize the pre-trained ViT-L/14 encoders from CLIP~\cite{radford2021learning}, which remain frozen during training. Unless stated otherwise, the watermark secret $\mathcal{S}$ is a randomly generated 16-bit binary vector. All training processes are conducted on 8 A100 NVIDIA GPUs with a batch size of $6$ per GPU.
All trainable parameters are optimized using the Adam optimizer with an initial learning rate of $1e-4$ and betas set to $(0.9, 0.999)$. We employ a step-based learning rate schedule with a gamma of $0.95$. For our composite loss function (Eq. \ref{loss_all}), the trade-off parameters $\{\lambda_1, \lambda_2, \lambda_3, \lambda_4\}$ were set to be $\{5, 5, 1, 1\}$, respectively. The model is trained for $10,000$ iterations. For the final evaluation, we use the checkpoint from the last training epoch.

\begin{table}[t]
    \centering
    \caption{
        Comparison of attribution performance on the WikiArt and ImageNet datasets. TokenTrace achieves the highest accuracy across both datasets, outperforming all passive and proactive baselines. ``Bit" represents bit accuracy (\%) and ``Att." represents attribution accuracy (\%). Passive baselines do not have Bit Accuracy as they do not retrieve a secret bit-string.
    }
    \vspace{-0.3cm}
    \resizebox{\linewidth}{!}{
    \begin{tabular}{l|c|cc|cc}
        \hline
        \multirow{2}{*}{\textbf{Method}} & \multirow{2}{*}{\textbf{Type}} & \multicolumn{2}{c|}{\textbf{WikiArt}} & \multicolumn{2}{c}{\textbf{ImageNet}} \\
        \cline{3-6}
        & & \textbf{Bit} $\uparrow$ & \textbf{Att} $\uparrow$ & \textbf{Bit} $\uparrow$ & \textbf{Att} $\uparrow$ \\
        \hline
        ALADIN~\cite{ruta2021aladin} & Passive & - & 18.58 & - & 5.55 \\
        CLIP~\cite{radford2021learning} & Passive & - & 52.60 & - & 42.61 \\
        Abc~\cite{wang2023evaluating} & Passive & - & 56.03 & - & 53.51 \\
        SSCD~\cite{pizzi2022self} & Passive & - & 45.34 & - & 25.50 \\
        EKILA~\cite{balan2023ekila} & Passive & - & 43.03 & - & 30.98 \\
        \hline
        ProMark~\cite{asnani2024promark} & Proactive & 93.14 & 87.19 & 90.56 & 87.30 \\
        CustomMark~\cite{asnani2025custom} & Proactive & 95.59 & 89.25 & 93.11 & 87.12 \\
        \hline
        \textbf{TokenTrace} & \textbf{Proactive} & \textbf{98.33} & \textbf{91.67} & \textbf{95.82} & \textbf{90.43} \\
        \hline
    \end{tabular}}
    \vspace{-0.5cm}
    \label{tab:main_results}
\end{table}

% \begin{table}[t]
%     \centering
%     % You can adjust the resizebox width. 0.8\linewidth is often a good size.
%     \resizebox{1\linewidth}{!}{%
%     \begin{tabular}{l|c c c}
%         \hline
%         \textbf{Method} & \textbf{Type} & \textbf{Bit Acc.} & \textbf{Attribution Acc.} \\
%         \hline
%         ALADIN~\cite{ruta2021aladin} & Passive & - & 18.58 \\
%         CLIP~\cite{radford2021learning} & Passive & - & 52.60 \\
%         AbC~\cite{wang2023evaluating} & Passive & - & 56.03 \\
%         SSCD~\cite{pizzi2022self} & Passive & - & 45.34 \\
%         EKILA~\cite{balan2023ekila} & Passive & - & 43.03 \\
%         \hline
%         ProMark~\cite{asnani2024promark} & Proactive & - & 87.19 \\
%         CustomMark~\cite{asnani2025custom} & Proactive & - & 89.25 \\
%         \hline
%         \textbf{TokenTrace} & \textbf{Proactive} & \textbf{98.33} & \textbf{91.67} \\
%         \hline
%     \end{tabular}
%     } % end resizebox
%     \vspace{-0.3cm}
%     \caption{%
%         Comparison of Attribution Accuracy on the WikiArt Dataset. TokenTrace, achieves the highest accuracy compared to both passive baselines and other proactive watermarking techniques in the task of artistic style attribution.
%     }
%     \label{tab:style_results}
% \end{table}

\begin{figure*}[t]
    \centering
    
    % --- Subfigure (a): Artistic Styles ---
    \begin{subfigure}[b]{0.49\linewidth}
        \centering
        \includegraphics[width=\linewidth]{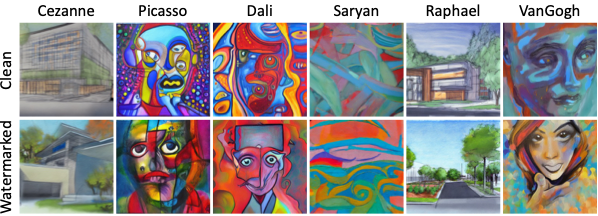}
        \caption{Qualitative Comparison for Artistic Style Concepts.}
        \label{fig:style_qualitative}
    \end{subfigure}
    \hfill % This adds horizontal space between the figures
    % --- Subfigure (b): Object Concepts ---
    \begin{subfigure}[b]{0.49\linewidth}
        \centering
        \includegraphics[width=\linewidth]{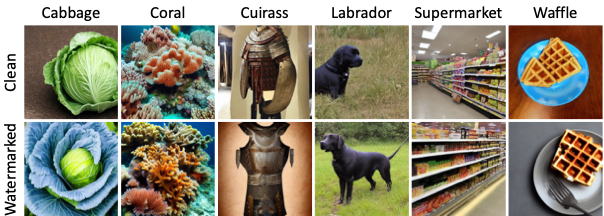}
        \caption{Qualitative Comparison for Single Object Concepts.}
        \label{fig:object_qualitative}
    \end{subfigure}
    \vspace{-0.3cm}
    \caption{
        Qualitative analysis of visual fidelity for watermarked images. (a) Results on abstract artistic style concepts from the WikiArt dataset. (b) Results on diverse object concepts from the ImageNet dataset. In both cases, the ``Watermarked" images are visually indistinguishable from the ``Clean" originals.
    }
    \vspace{-0.5cm}
    \label{fig:qualitative_all}
\end{figure*}

% \begin{figure*}
%     \centering
%     \includegraphics[width=1\linewidth]{figs/multi_concept_performance.png}
%     \vspace{-0.8cm}
%     \caption{
%     Performance comparison for multi-concept attribution. The left plot shows results for attributing 2 customized concepts (one object and one style), while the right plot shows results for 4 general concepts. TokenTrace, consistently outperforms the CustomMark baseline in both Bit Accuracy and Attribution Accuracy. Performance is further improved by applying prompt weighting (TokenTraceP).}
%     \vspace{-0.5cm}
%     \label{fig:multi}
% \end{figure*}

\begin{table}[t]
\centering
\caption{Performance comparison for multi-concept attribution. The left plot shows results for attributing 2 customized concepts (one object and one style), while the right plot shows results for 4 general concepts. TokenTrace, consistently outperforms the CustomMark baseline in both bit accuracy, represented as ``Bit", and attribution accuracy, represented as ``Attrbution". Performance is further improved by applying prompt weighting (TokenTraceP). All accuracy metrics are in percent (\%).}
\vspace{-0.3cm}
\label{tab:multi}
\resizebox{1\linewidth}{!}{
\begin{tabular}{l cc cc}
\toprule
& \multicolumn{2}{c}{\textbf{Customized}} & \multicolumn{2}{c}{\textbf{General}} \\
\cmidrule(lr){2-3} \cmidrule(lr){4-5}
\textbf{Method} & Bit $\uparrow$ & Attribution $\uparrow$ & Bit $\uparrow$ & Attribution $\uparrow$ \\
\midrule
CustomMark & 92.47 & 85.14 & 78.93 & 72.78 \\
TokenTrace & 94.15 & 88.62 & 85.41 & 81.57 \\
TokenTraceP & \textbf{96.83} & \textbf{90.53} & \textbf{90.33} & \textbf{86.08} \\
\bottomrule
\end{tabular}}
\end{table}

\begin{figure}[t]
    \centering
    \includegraphics[width=1\linewidth]{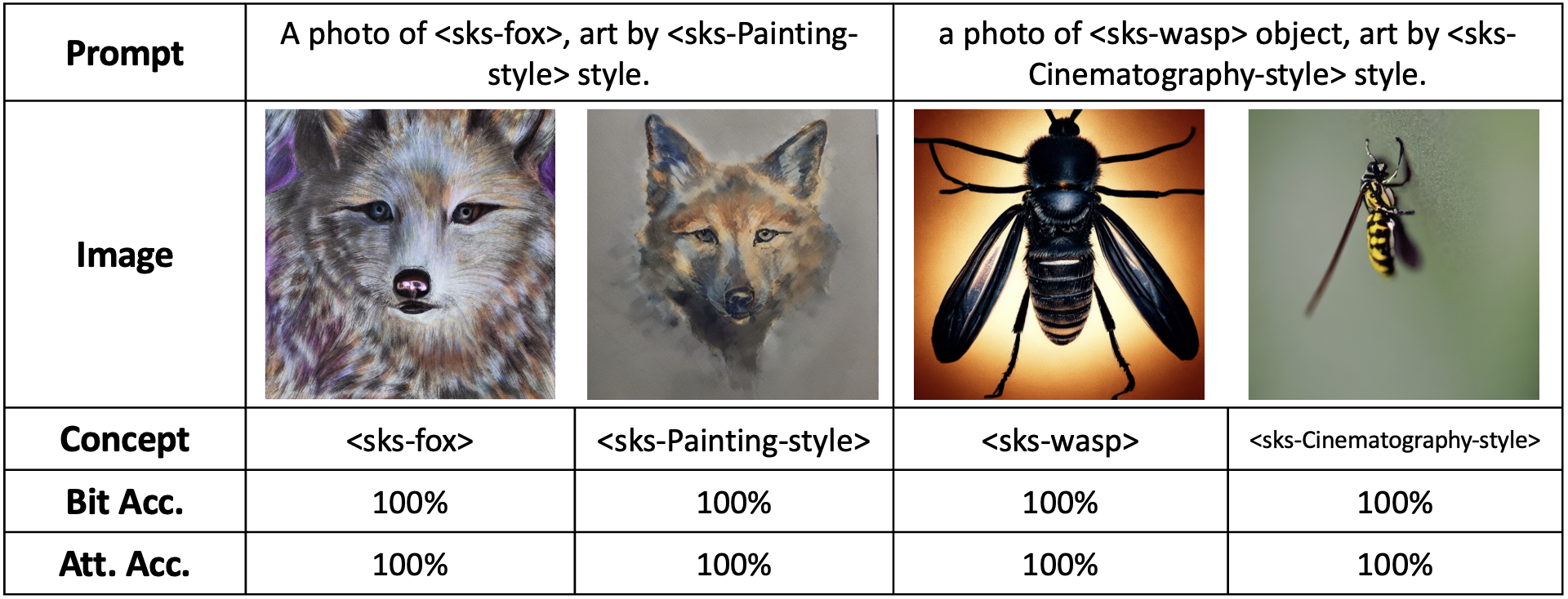}
    \vspace{-0.6cm}
    \caption{
        Qualitative example of multi-customized concept prediction. We generate two images using a single prompt containing multiple watermarked concepts, and report the average bit accuracy and average attribution accuracy for each prompt.
    }
    \vspace{-0.3cm}
    \label{fig:qualitative_multi_customized}
\end{figure}

% \begin{figure}[t]
%     \centering
%     % Adjust the width as needed. 0.8\linewidth often looks good.
%     \includegraphics[width=1\linewidth]{figs/qua_wikiart.png} % Your figure file
%     \vspace{-0.6cm}
%     \caption{%
%         Qualitative Comparison for Artistic Style Concepts. The image pairs show that our watermarking process preserves the complex textures, colors, and overall composition of various artistic styles, demonstrating high visual fidelity.
%     }
%     \label{fig:style_qualitative}
% \end{figure}

\subsection{Results and Analysis}

\paragraph{Single Style Concept Attribution.} 
We first evaluate our method's ability to attribute abstract concepts, such as artistic styles on the WikiArt~\cite{tan2018improved} dataset. We compared TokenTrace  against a series of both passive and proactive attribution baselines. The attribution accuracy and the bit prediction accuracy was calculated based on $100$ generated images. As shown in Table~\ref{tab:main_results}, the results clearly demonstrate the superiority of our semantic watermarking approach for this complex task.
\begin{itemize}
    \item Firstly, TokenTrace achieves a state-of-the-art Attribution Accuracy of $91.67\%$, significantly outperforming all other baselines.
    \item Secondly, TokenTrace surpasses all passive attribution methods by a large margin. For instance, it is substantially more accurate than standard CLIP retrieval ($52.60\%$), confirming that simple similarity is insufficient for robust attribution.
    \item Thirdly, our method also outperforms other proactive watermarking techniques. It shows a clear improvement over both the pixel-based ProMark ($87.19\%$) and the latent-based CustomMark ($89.25\%$).
\end{itemize}

The qualitative results in Figure~\ref{fig:style_qualitative} further validate our method's performance. The ``Watermarked" images successfully retain the intricate details and overall aesthetic of the original artistic styles. The visual fidelity between the ``Clean" and ``Watermarked" pairs is high. This confirms that watermarks inserted by TokenTrace is imperceptible, a crucial requirement for any practical attribution system.

\vspace{-0.3cm}
\paragraph{Single Object Concept Attribution.} 
We further assess evaluate the scalability and performance of TokenTrace on discrete objects, we conducted a large-scale experiment on $1000$ object concepts sourced from the ImageNet~\cite{deng2009imagenet} dataset.
Similar from the style attribution task, the results presented in Table~\ref{tab:main_results} clearly demonstrates the superiority of our approach. TokenTrace achieves a state-of-the-art attribution accuracy of $90.43\%$ and a bit accuracy of $95.82\%$. It significantly outperforms all passive methods and shows a clear performance gain over the other proactive baselines, ProMark ($87.30\%$) and CustomMark ($87.12\%$). 
The qualitative results in Figure~\ref{fig:object_qualitative} further confirm the imperceptibility of our watermark. Across diverse concepts, the ``Watermarked" images are visually indistinguishable from their ``Clean" images, demonstrating that TokenTrace achieves robust attribution while maintaining high visual fidelity.

\vspace{-0.3cm}
\paragraph{Multi-Concept Attribution.} A key advantage of our framework is attributing multiple concepts within a single image. We first evaluated this on two customized concepts on the self-created concept pool of 20 concepts (10 objects and 10 artistic styles). As shown in Table~\ref{tab:multi}, TokenTrace ($88.62\%$) outperformed the CustomMark baseline ($85.14\%$) upon the comparison of attribution accuracy, and an enhanced version, TokenTraceP (with prompt weighting)~\footnote{Details about TokenTraceP can be found in supplementary material.}, further boosted this accuracy to $90.53\%$. We then tested a more complex four-concept scenario. Because composing multiple customized concepts degrades image quality, we constructed a new benchmark using ChatGPT~\cite{brown2020language} to generate a diverse set of prompts involving four distinguished concepts. The results in Table~\ref{tab:multi} show our advantage becomes even more pronounced: TokenTrace ($81.57\%$) dramatically outperformed CustomMark ($72.78\%$) upon attribution accuracy, while TokenTraceP again achieved the highest accuracy ($86.08\%$).

\begin{figure}[t]
    \centering
    \includegraphics[width=1\linewidth]{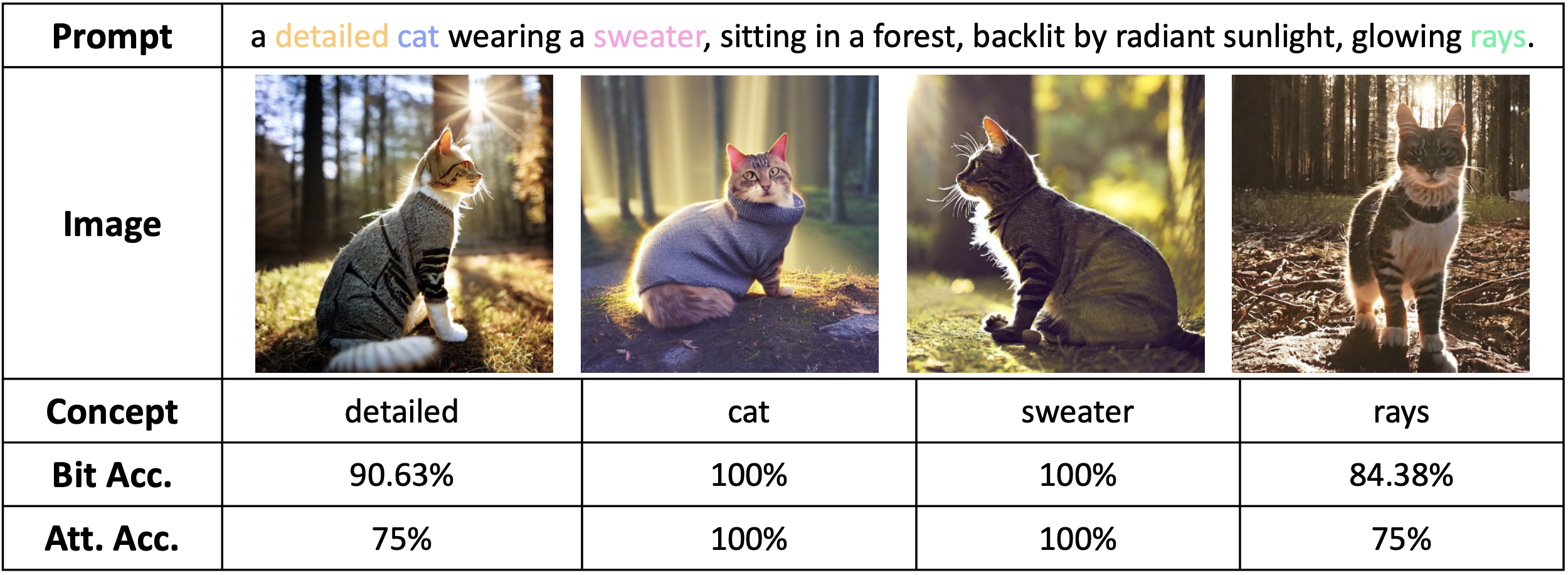}
    \vspace{-0.6cm}
    \caption{
        Qualitative example of multi-general concept prediction. We generate four images using a single prompt containing multiple watermarked concepts, and report the average bit accuracy and average attribution accuracy.
    }
    \vspace{-0.3cm}
    \label{fig:qualitative_multi_general}
\end{figure}

Finally, Figure~\ref{fig:qualitative_multi_customized} and Figure~\ref{fig:qualitative_multi_general} provides qualitative proof of this disentanglement. Specifically, from a single image generated with four watermarked concepts (e.g., ``detailed cat wearing a sweater... glowing rays"), our query-based module successfully retrieves the correct, independent secret for each concept, retrieving 'cat' and 'sweater' with $100\%$ accuracy and 'detailed' and 'rays' with high accuracy. This confirms our query-based mechanism solves concept overlap and scales to complex prompts. More qualitative examples can be found in the supplementary material.

\begin{figure}[t]
    \centering
    \includegraphics[width=1\linewidth]{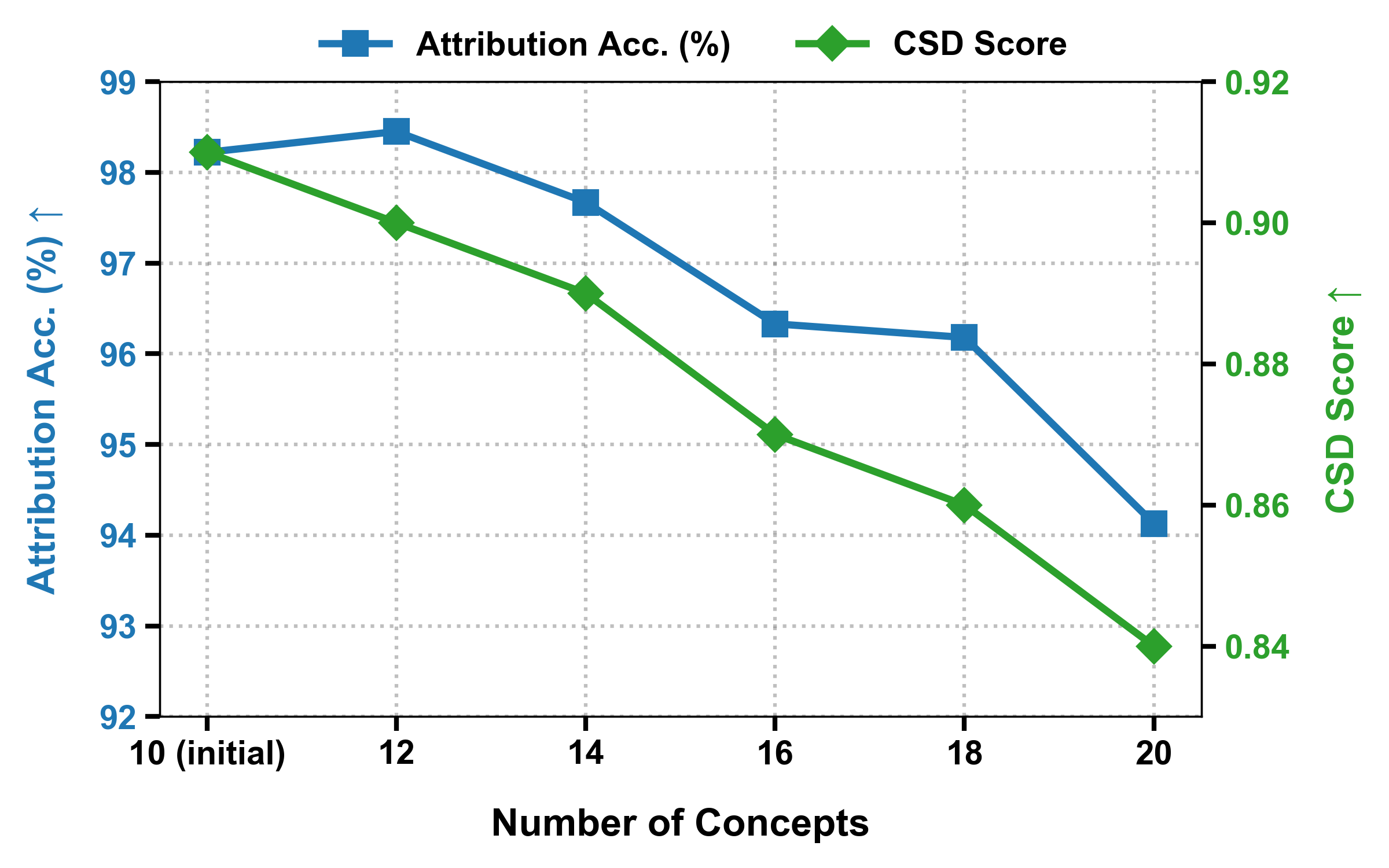}
    \vspace{-0.8cm}
    \caption{Performance on incremental concept learning. The plot shows the attribution accuracy (left y-axis, blue squares) and CSD score (right y-axis, green diamonds) as new concepts are sequentially added to a pre-trained model. The results demonstrate that both attribution accuracy and visual fidelity remain high, showing the method's efficiency in real-world applications.}
    \vspace{-0.5cm}
    \label{fig:seq}
\end{figure}

\vspace{-0.3cm}
\paragraph{Sequential Learning.}
To evaluate our framework's ability to add new concepts without full retraining, we perform a sequential learning experiment on the WikiArt dataset. We first train an initial TokenTrace model on $10$ style concepts, then sequentially introduce $2$ new concepts at a time up to $20$. Critically, we do not retrain from scratch but instead finetune the existing model with only 10\% additional iterations per step, reflecting a real-world scenario. The results in Figure~\ref{fig:seq} demonstrate high efficiency and strong resistance to catastrophic forgetting. Both Attribution Accuracy (starting at $98.22\%$ and ending at $94.13\%$) and CSD Score (starting at $0.91$ and ending at $0.84$) remain high, confirming that our parameter-efficient design can be dynamically updated, making it a scalable and practical solution.

\subsection{Ablation Studies}

\paragraph{Ablation on Loss Components.}
Our composite loss function ($\mathcal{L}_{\text{total}}$), as defined in Eq.~\ref{loss_all}, is designed to jointly optimize for retrieval accuracy and visual fidelity. To demonstrate the contribution of each loss term, we train our model from scratch with different components of the objective function removed. The full model (``All") is trained with all loss components, including $\mathcal{L}_{\text{BCE}}$, $\mathcal{L}_{\text{CSD}}$, $\mathcal{L}_{\text{L2}}$, and $\mathcal{L}_{\text{reg}}$. 
As shown in Table~\ref{tab:ablation_loss}, our complete model (``All") achieves the best performance. Removing the CSD loss (``No CSD") causes the most significant performance drop. This finding confirms that the $\mathcal{L}_{\text{CSD}}$ term is critical for maintaining the high-level semantic and style consistency required for accurate retrieval. Removing the image-space L2 loss (``No L2 (Image)") also leads to a notable decrease in performance, with attribution accuracy dropping to $86.37\%$ and the CSD Score falling to $0.73$. The latent-space L2 (``No L2 (Latent)") also contributes, with its removal dropping accuracy to $88.52\%$. 

\begin{table}[t]
    \centering
    \caption{
        Ablation study for the objective function on the WikiArt dataset. Our full model (``All") significantly outperforms variants where individual loss components (CSD, Latent L2, Image L2) are removed, demonstrating that each component is necessary for achieving both high attribution accuracy and visual fidelity.
    }
    \vspace{-0.3cm}
    \resizebox{.9\columnwidth}{!}{%
    % --- Start of the SINGLE table ---
    \begin{tabular}{l|c c}
        \hline
        \multicolumn{3}{c}{\textbf{Attribution Performance (\%)}} \\
        \hline
        \textbf{Method} & \textbf{Bit Acc.} $\uparrow$ & \textbf{Attribution Acc.} $\uparrow$ \\
        \hline
        No CSD & 91.81 & 83.75 \\
        No L2 (latent) & 96.03 & 88.52 \\
        No L2 (Image) & 93.65 & 86.37 \\
        \textbf{All} & \textbf{98.33} & \textbf{91.67} \\
        \hline\hline  % --- This creates the strong dividing line ---
        \multicolumn{3}{c}{\textbf{Image Quality}} \\
        \hline
        \textbf{Method} & \textbf{CLIP Score} $\uparrow$ & \textbf{CSD Score} $\uparrow$ \\
        \hline
        No CSD & 0.73 & 0.65 \\
        No L2 (latent) & 0.82 & 0.76 \\
        No L2 (Image) & 0.81 & 0.73 \\
        \textbf{All} & \textbf{0.87} & \textbf{0.82} \\
        \hline
    \end{tabular}
    % --- End of the SINGLE table ---
    } % end resizebox
    \vspace{-0.3cm}
    \label{tab:ablation_loss}
\end{table}

\begin{figure}
    \centering
    \includegraphics[width=1\linewidth]{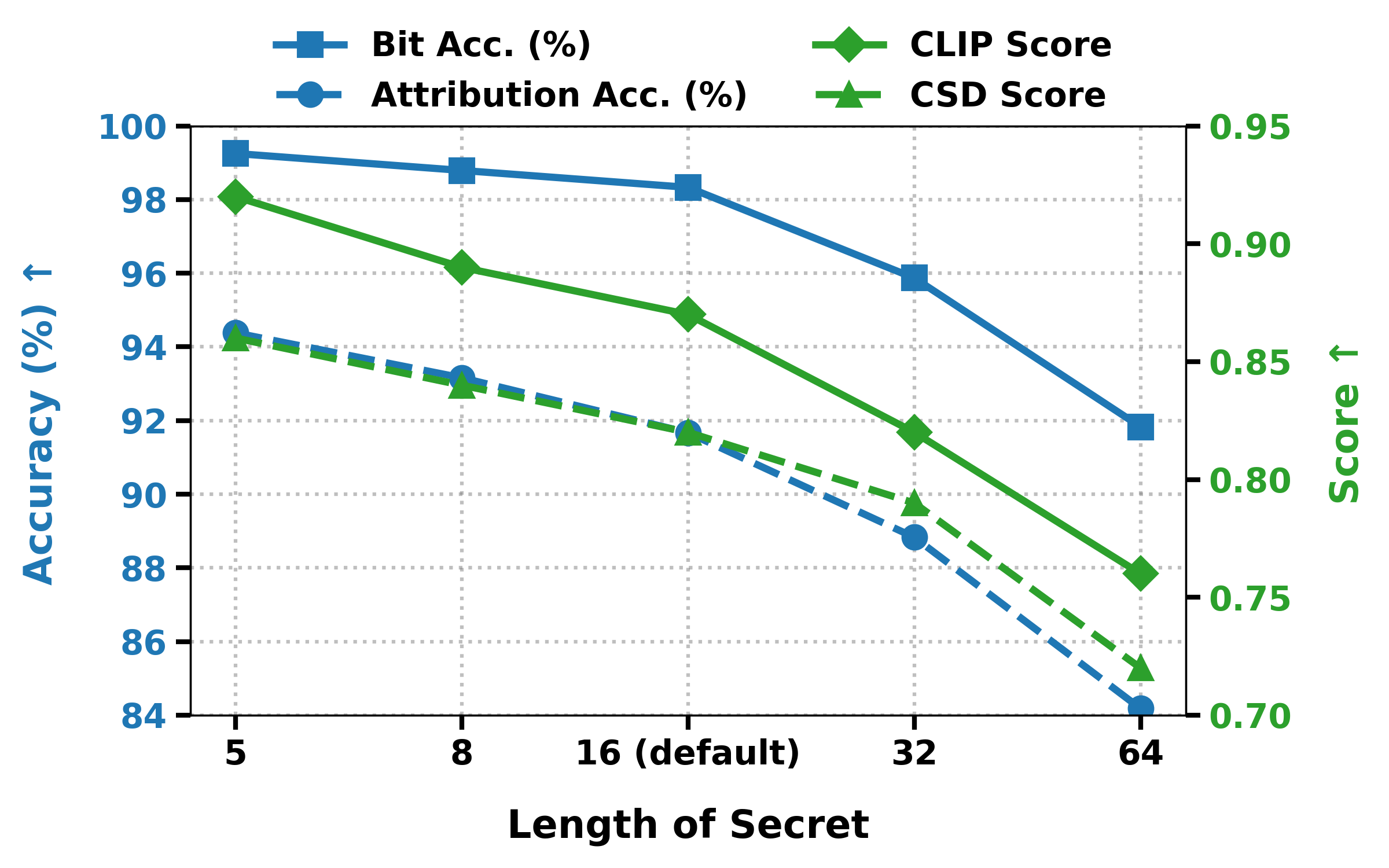}
    \vspace{-0.8cm}
    \caption{Ablation study for the length of the bit secret on the WikiArt dataset. Performance shows a graceful trade-off between capacity and accuracy. Our 16-bit default provides the best balance of high attribution accuracy and visual fidelity.}
    \label{fig:ablation_bit_length}
\end{figure}

\vspace{-0.3cm}
\paragraph{Ablation on Bit Secret Length.}
We next investigate the impact of the length of the bit secret $\mathcal{S}$, on both attribution accuracy and visual fidelity. We trained separate models on the WikiArt dataset using secret lengths of 5, 8, 16 (our default), 32, and 64 bits. As shown in Figure~\ref{fig:ablation_bit_length}, we observe a clear and expected trade-off between information capacity and performance. Specifically, The model achieves the highest attribution accuracy ($94.38\%$) with a 5-bit secret, as this smaller signal is easiest to embed and retrieve. As the bit length increases, all metrics show a graceful degradation. When moving from our 16-bit default to a 64-bit secret, the attribution accuracy drops from $91.67\%$ to $84.18\%$, and the CSD score falls from $0.82$ to $0.72$. This trend is logical, as embedding a larger and more complex signal (64 bits) into the image is an inherently harder optimization problem and is more likely to interfere with the image's original semantic content, thus slightly reducing visual fidelity. These findings validate our choice of 16 bits as the default, as it provides an balance, offering a sufficiently large space for unique signatures while maintaining both high attribution accuracy and high visual quality.

% \begin{table}[h]
%     \centering
%     \resizebox{\columnwidth}{!}{%
%     \begin{tabular}{l|c c c c}
%         \hline
%         \textbf{Method} & \textbf{Bit Acc. (\%)} $\uparrow$ & \textbf{Attribution Acc. (\%)} $\uparrow$ & \textbf{CLIP Score} $\uparrow$ & \textbf{CSD Score} $\uparrow$ \\
%         \hline
%         5 & 99.25 & 94.38 & 0.92 & 0.86 \\
%         8 & 98.79 & 93.16 & 0.89 & 0.84 \\
%         \textbf{16 (default)} & \textbf{98.33} & \textbf{91.67} & \textbf{0.87} & \textbf{0.82} \\
%         32 & 95.87 & 88.83 & 0.82 & 0.79 \\
%         64 & 91.83 & 84.18 & 0.76 & 0.72 \\
%         \hline
%     \end{tabular}
%     } % end resizebox
%     \caption{
%         Ablation study for the length of the bit secret on the WikiArt dataset. Performance shows a graceful trade-off between capacity and accuracy. Our 16-bit default provides the best balance of high attribution accuracy and visual fidelity.
%     }
%     \label{tab:ablation_bit_length}
% \end{table}

\begin{figure}
    \centering
    \includegraphics[width=1\linewidth]{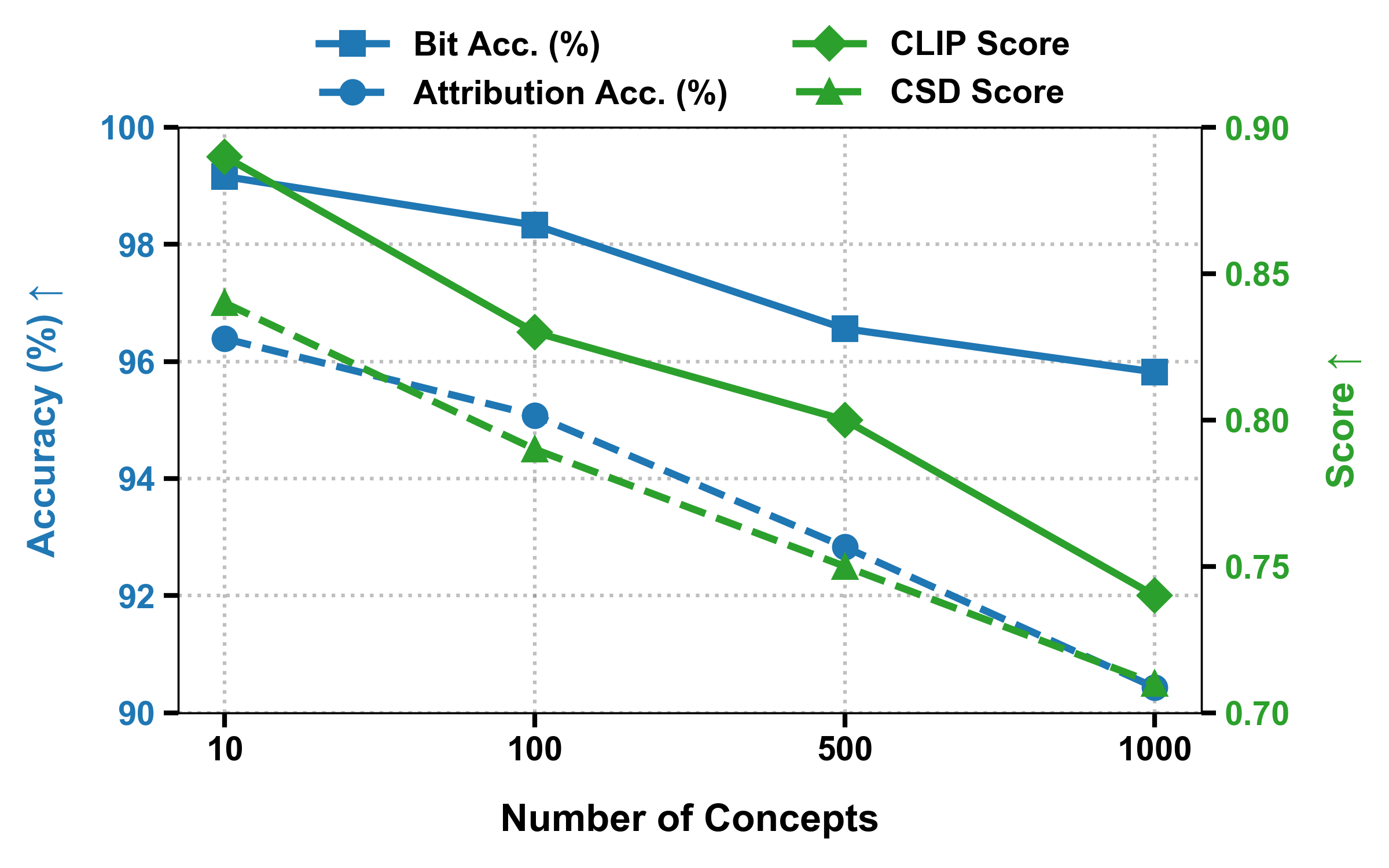}
    \vspace{-0.8cm}
    \caption{Ablation study for the number of concepts on the ImageNet dataset. Performance remains high and degrades gracefully as the number of concepts scales from $10$ to $1000$, demonstrating the high scalability of our method.}
    \label{fig:ablation_num_concepts}
\end{figure}

\vspace{-0.3cm}
\paragraph{Ablation on the Number of Concepts.}
We next evaluate our framework's scalability by training models on the ImageNet dataset with concept pools of increasing size: 10, 100, 500, and 1000 concepts. The results in Figure~\ref{fig:ablation_num_concepts} demonstrate a very high level of performance and graceful degradation. As the concept library scales from 10 to 1000 concepts, the attribution accuracy remains exceptionally high, dropping only ~6 percentage points ($96.39\% \rightarrow 90.43\%$). Similarly, bit accuracy shows strong resilience ($99.16\% \rightarrow 95.82\%$), and visual fidelity scales well, with the CSD score dropping only from $0.84$ to $0.71$. These findings demonstrate that TokenTrace is highly scalable and can robustly handle a large library of distinct concepts, confirming its practicality for real-world applications.

% \begin{table}[h]
%     \centering
%     \resizebox{\columnwidth}{!}{%
%     \begin{tabular}{l|c c c c}
%         \hline
%         \textbf{Method} & \textbf{Bit Acc. (\%)} $\uparrow$ & \textbf{Attribution Acc. (\%)} $\uparrow$ & \textbf{CLIP Score} $\uparrow$ & \textbf{CSD Score} $\uparrow$ \\
%         \hline
%         10 & 99.16 & 96.39 & 0.89 & 0.84 \\
%         100 & 98.33 & 95.08 & 0.83 & 0.79 \\
%         500 & 96.56 & 92.83 & 0.80 & 0.75 \\
%         1000 & 95.82 & 90.43 & 0.74 & 0.71 \\
%         \hline
%     \end{tabular}
%     } % end resizebox
%     \caption{
%         Ablation study for the number of concepts on the ImageNet dataset. Performance remains high and degrades gracefully as the number of concepts scales from 10 to 1000, demonstrating the high scalability of our method.
%     }
%     \label{tab:ablation_num_concepts}
% \end{table}

\vspace{-0.3cm}
\paragraph{Robustness to Image Distortions.}
Finally, to validate our watermark's resilience, we conducted a robustness analysis by applying common image distortions to watermarked WikiArt images before retrieval, simulating real-world re-encoding and editing. The results in Table~\ref{tab:ablation_distortions} demonstrate that our dual-conditioning strategy creates a highly robust watermark. The method shows exceptional resilience, maintaining over $90.04\%$ attribution accuracy against Rotation and over $87\%$ against both JPEG compression and a targeted Adversarial Attack. Performance also degrades gracefully for more severe transformations, with CropAndResize ($86.57\%$), GaussianBlur ($84.81\%$), ColorJitter ($83.22\%$), and GaussianNoise ($82.94\%$) all retaining over $82\%$ accuracy. These results confirm that by embedding the signature in both the semantic and latent domains, our watermark is a deeply integrated signal, allowing for reliable attribution even after significant image alterations.

\begin{table}[h]
    \centering
    \caption{
        Ablation study for different types of distortions on the WikiArt dataset. Our method demonstrates high robustness, maintaining strong attribution accuracy. The detailed description of ``Adversarial Attack" can be found in supplementary material.
    }
    \vspace{-0.2cm}
    \resizebox{\columnwidth}{!}{%
    \begin{tabular}{l|c c}
        \hline
        \textbf{Method} & \textbf{Bit Acc.} $\uparrow$ & \textbf{Attribution Acc.} $\uparrow$ \\
        \hline
        No distortion & 98.33 & 91.67 \\
        \hline
        JPEG & 94.68 & 88.20 \\
        Rotation & 96.21 & 90.04 \\
        CropAndResize & 93.28 & 86.57 \\
        GaussianBlur & 91.32 & 84.81 \\
        GaussianNoise & 89.18 & 82.94 \\
        ColorJitter & 89.62 & 83.22 \\
        Sharpness & 92.71 & 86.54 \\
        Adversarial Attack~\cite{zhao2024invisible} & 94.08 & 87.17 \\
        \hline
    \end{tabular}
    } % end resizebox
    \vspace{-0.5cm}
    \label{tab:ablation_distortions}
\end{table}

\section{Conclusion}
\label{sec:conclusion}

In this paper, we introduced TokenTrace, a novel proactive watermarking framework that solves the challenging task of multi-concept attribution. By embedding a secret into both the textual semantic and latent domains, our method creates a robust signature that avoids the spatial overlap problem plaguing pixel-based methods. Our key contribution is the query-based TokenTrace module, which can successfully disentangle and independently retrieve signatures for multiple, composed concepts from a single image. Extensive experiments demonstrate that TokenTrace significantly outperforms state-of-the-art baselines on single-concept (style and object) and multi-concept tasks, all while maintaining high visual fidelity and robustness to distortions.
{
    \small
    \bibliographystyle{ieeenat_fullname}
    \bibliography{main}
}

% WARNING: do not forget to delete the supplementary pages from your submission 

\end{document}